\documentclass{article}  %
\usepackage{iclr2026_conference,times}

\usepackage{amsmath,amsfonts,bm}

\def\eqref#1{equation~\ref{#1}}

\def\1{\bm{1}}

\DeclareMathAlphabet{\mathsfit}{\encodingdefault}{\sfdefault}{m}{sl}
\SetMathAlphabet{\mathsfit}{bold}{\encodingdefault}{\sfdefault}{bx}{n}

\usepackage{capt-of, eucal}
\usepackage{xspace}
\usepackage{xcolor}
\usepackage{enumitem}
\usepackage{amsmath,amsthm,amssymb,amsfonts,dsfont,pifont,bm,bbm,mathrsfs,mathtools,nicefrac,extarrows,relsize}
\usepackage{algorithm,algpseudocode,listings}
\usepackage{booktabs,multirow,adjustbox,diagbox,threeparttable,tabularray,setspace}
\usepackage{wrapfig}
\usepackage{graphicx}
\usepackage{subcaption}
\usepackage{tabularx}
\usepackage{array}
\usepackage{makecell}
\usepackage{longtable} 
\usepackage{paracol}
\usepackage{float}
\usepackage{fontawesome5}
\usepackage[tableposition=top]{caption}
\usepackage{colortbl}
\usepackage{color}
\usepackage{multicol}
\usepackage{lipsum}

\definecolor{citeblue}{rgb}{0.21,0.49,0.74}
\usepackage[pagebackref=false,breaklinks,colorlinks,citecolor=citeblue,bookmarks=false]{hyperref}
\usepackage{wrapfig}
\usepackage[capitalize]{cleveref}  %

\crefname{section}{Sec.}{Secs.}
\Crefname{section}{Section}{Sections}
\crefname{appendix}{Appendix}{Appendices}
\Crefname{appendix}{Appendix}{Appendices}
\crefname{table}{Table}{Tables}
\Crefname{table}{Table}{Tables}
\crefname{figure}{Fig.}{Figs.}
\Crefname{figure}{Figure}{Figures}
\crefname{equation}{Eq.}{Eqs.}
\Crefname{equation}{Equation}{Equations}
\crefname{theorem}{Thm.}{Thms.}
\Crefname{theorem}{Theorem}{Theorems}
\crefname{lemma}{Lem.}{Lems.}
\Crefname{lemma}{Lemma}{Lemmas}
\crefname{remark}{Rem.}{Rems.}
\Crefname{remark}{Remark}{Remarks}
\crefname{corollary}{Cor.}{Cors.}
\Crefname{corollary}{Corollary}{Corollaries}
\crefname{algorithm}{Alg.}{Algs.}
\Crefname{algorithm}{Algorithm}{Algorithms}
\definecolor{cellred}{RGB}{213, 123, 101}
\definecolor{cellgreen}{RGB}{0, 205, 0}
\definecolor{cellblue}{RGB}{54, 125, 189}
\definecolor{codegreen}{rgb}{0,0.6,0}
\definecolor{codegray}{rgb}{0.5,0.5,0.5}
\definecolor{codepurple}{rgb}{0.58,0,0.82}
\definecolor{backcolour}{rgb}{1.0,1.0,1.0}
\lstdefinestyle{mystyle}{
    backgroundcolor=\color{backcolour},
    commentstyle=\color{codegreen},
    keywordstyle=\color{magenta},
    numberstyle=\tiny\color{codegray},
    stringstyle=\color{codepurple},
    basicstyle=\ttfamily\scriptsize,
    breakatwhitespace=false,
    breaklines=true,
    captionpos=b,
    keepspaces=true,
    numbers=left,
    numbersep=5pt,
    showspaces=false,
    showstringspaces=false,
    showtabs=false,
    tabsize=2
}
\usepackage[most,skins,theorems]{tcolorbox}
\tcbset{
  aibox/.style={
    width=\linewidth,
    top=8pt,
    bottom=4pt,
    colback=blue!6!white,
    colframe=black,
    colbacktitle=black,
    enhanced,
    center,
    attach boxed title to top left={yshift=-0.1in,xshift=0.15in},
    boxed title style={boxrule=0pt,colframe=white,},
  }
}
\newtcolorbox{AIbox}[2][]{aibox,title=#2,#1}
\usepackage{amsmath} %
\usepackage{amssymb} %
\usepackage{multirow}

\usepackage{rotating}  %

\newcolumntype{C}[1]{>{\centering\arraybackslash}p{#1}}
\newcolumntype{L}[1]{>{\arraybackslash}p{#1}}

\definecolor{demphcolor}{gray}{.2}

\definecolor{demphcolorinline}{gray}{.3}

\definecolor{demphcolor1}{gray}{.6}

\newcommand{\tocite}[1]{{\color{red} [TO CITE]}}

\title{LAPO: Internalizing Reasoning Efficiency via Length-Adaptive Policy Optimization}

\author{%
  \textbf{Xingyu Wu}$^{1}$,
  ~~
  \textbf{Yuchen Yan}$^{1}$,
  ~~ 
  \textbf{Shangke Lyu}$^{1}$,
  ~~
  \textbf{Linjuan Wu}$^{1}$,
  ~~
  \textbf{Yiwen Qiu}$^{1}$,
  ~~
  \\
  \textbf{Yongliang Shen}$^{1}$,
  ~~
    \textbf{Weiming Lu}$^{1}$,
  ~~
  \textbf{Jian Shao}$^{1}$,
  ~~ 
  \textbf{Jun Xiao}$^{1}$,
  ~~ 
  \textbf{Yueting Zhuang}$^{1}$ \\
  $^1$Zhejiang University\\
  \texttt{\{wuxingyu, syl\}@zju.edu.cn} \\
  \begin{tabular}{@{}ll@{}}
  \vspace{0.1cm} \\
    \faGithub\ GitHub: & \href{https://github.com/zju-real/lapo}{\texttt{\textcolor{cyan}{https://github.com/zju-real/lapo}}} \\
    \faGlobe\ Project: & \href{https://zju-real.github.io/lapo}{\texttt{\textcolor{cyan}{https://zju-real.github.io/lapo}}}
  \end{tabular}
}

\iclrfinalcopy %

\begin{document}

\maketitle

\begin{abstract}

Large reasoning models have achieved remarkable performance through extended chain-of-thought sequences, yet this computational freedom leads to excessive token generation even for simple problems. We present Length-Adaptive Policy Optimization (LAPO), a novel framework that transforms reasoning length control from an external constraint into an intrinsic model capability. Unlike existing approaches that impose rigid limits or rely on post-hoc interventions, LAPO enables models to internalize an understanding of appropriate reasoning depth through a two-stage reinforcement learning process. In the first stage, models learn natural reasoning patterns by discovering the statistical distribution of successful solution lengths. The second stage leverages these patterns as meta-cognitive guidance, embedding them directly within the model's reasoning context to ensure inference-time flexibility. Experiments on mathematical reasoning benchmarks demonstrate that LAPO reduces token usage by up to 40.9\% while improving accuracy by 2.3\%. Our analysis reveals that models trained with LAPO develop emergent abilities to allocate computational resources based on problem complexity, achieving efficient reasoning without sacrificing quality.

\end{abstract}

\section{Introduction}
Recent advances in large reasoning models have demonstrated remarkable capabilities through extended chain-of-thoughts \citep{30, 27, 26}. However, this computational freedom comes with a critical limitation: these models often generate excessively verbose reasoning chains regardless of problem complexity, leading to significant computational overhead, a phenomenon known as "overthinking" \citep{23}. 
While beneficial for complex problems, this verbosity introduces substantial inefficiencies, making practical deployment challenging.

Existing approaches to address this challenge fall into three main categories, each with inherent limitations. Direct length reduction methods either rely on reward design \citep{34, 35} that can cause over-shortening and accuracy degradation, or impose hard length constraints \citep{14, 29} that lack adaptability across problem types. Dynamic early-stopping approaches \citep{33, 6} make real-time termination decisions but often truncate mid-reasoning, disrupting the thinking process. Adaptive thinking methods \citep{38, 37, 36} enable models to switch between thinking and non-thinking modes but operate at a coarse granularity.

The fundamental limitation of these approaches lies in their treatment of length control as an external constraint imposed upon the reasoning process. This paradigm inherently conflicts with the nature of mathematical reasoning, where each problem possesses its own intrinsic complexity that naturally determines the required reasoning depth. Current methods fail to recognize that when models successfully solve problems, they naturally converge to certain reasoning lengths that reflect this intrinsic complexity. The challenge is not to impose arbitrary limits, but to help models discover and internalize these natural reasoning patterns.

We propose a paradigm shift: instead of constraining reasoning through external mechanisms, we enable models to learn from their own successful reasoning patterns and develop an internal sense of appropriate reasoning depth. Our key insight is that the distribution of reasoning lengths in correct solutions contains valuable information about how much thinking each problem genuinely requires. By capturing these patterns during training and teaching models to anticipate the appropriate reasoning budget before they begin solving, we can transform length control from an external limitation into an intrinsic capability.

\begin{figure}[t!]
\centering
\includegraphics[width=\linewidth]{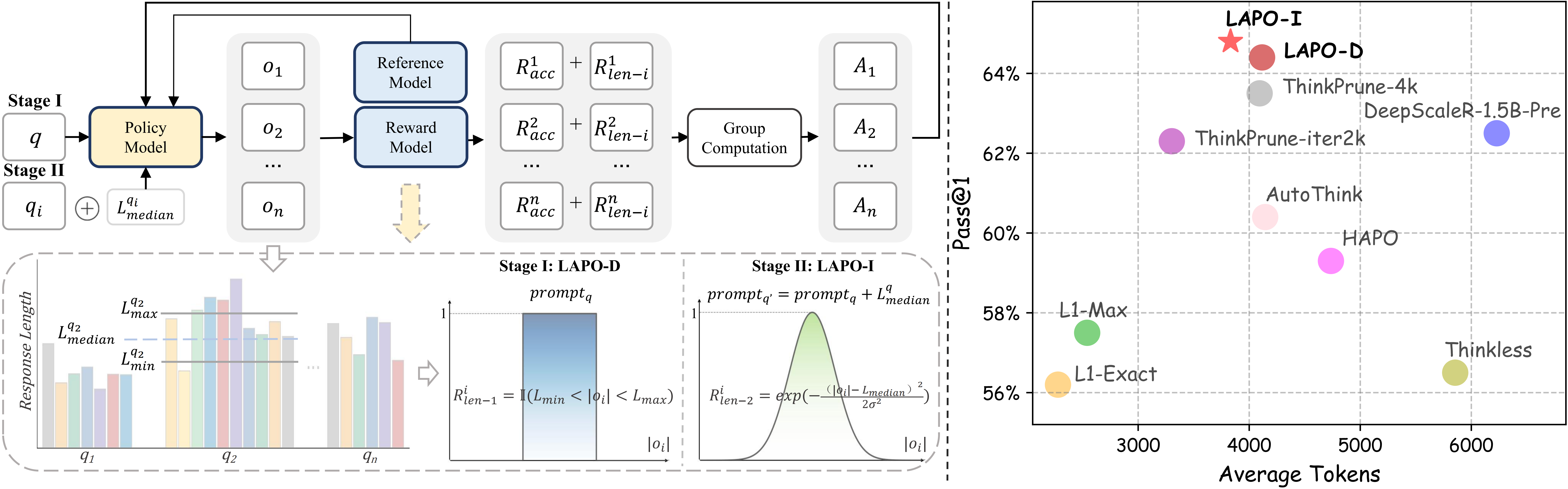}
\caption{Overview of Length-Adaptive Policy Optimization (LAPO) and its superior performance. The LAPO framework (left) trains a model in two stages: first discovering natural reasoning lengths, then internalizing them as self-proposed budgets. This process enables our models (LAPO-I) to achieve a state-of-the-art balance between accuracy and efficiency (right), surpassing existing methods by operating in the desirable top-left region of the performance plot.}
\label{fig-0}
\end{figure}

We introduce Length-Adaptive Policy Optimization (LAPO), a two-stage reinforcement learning framework that progressively builds this adaptive capability. In the first stage, we design length-aware rewards that encourage efficiency while maintaining accuracy. During this process, we collect statistical patterns from successful solutions, specifically focusing on the reasonable length range where most correct answers naturally fall. This reveals the inherent reasoning requirements of each problem without imposing artificial constraints. In the second stage, we leverage these discovered patterns to provide explicit guidance to the model. By incorporating target length information directly into the problem prompt, we enable models to plan their reasoning trajectory before beginning the solution process. Crucially, to ensure models can reason adaptively without requiring predefined lengths at inference time, we embed the length constraint as a self-declarative statement immediately after the \texttt{<think>} token. This technique reframes the budget not as an external command, but as part of the model’s own internal reasoning plan. By learning to generate a solution that aligns with its self-proposed budget, the model is incentivized to internalize the link between problem complexity and resource allocation, allowing it to reason flexibly and efficiently when deployed.

LAPO fundamentally differs from existing approaches by recognizing that efficient reasoning requires understanding problem-specific computational needs rather than following rigid rules. Our two-stage design enables a natural progression: models first learn what constitutes appropriate reasoning depth through experience, then develop the ability to anticipate these requirements proactively. This approach mirrors how human experts develop intuition about problem complexity, allocating mental effort proportionally to task demands.

Extensive experiments validate the effectiveness of our approach. LAPO achieves remarkable efficiency gains, reducing token usage by up to 40.9\% while simultaneously improving accuracy by 2.3\% on mathematical reasoning benchmarks (see Figure \ref{fig-0}). Our analysis reveals that this improvement stems from the model's ability to distinguish between problems requiring elaborate derivations versus those needing only brief calculations. The training dynamics demonstrate smooth convergence in both stages, with models maintaining stable accuracy even as they learn increasingly precise length control. These results confirm that when models learn from their own successful patterns rather than arbitrary constraints, they develop more robust and efficient reasoning strategies.

Our main contributions are:
\begin{itemize}
\item We propose LAPO, a novel two-stage reinforcement learning framework that transforms length control from an external constraint into an intrinsic reasoning capability, enabling models to adaptively allocate computational resources based on problem complexity.
\item We introduce a training methodology that combines statistical analysis of successful reasoning patterns with contextual length guidance embedded within the reasoning process, allowing models to internalize length-adaptive behaviors while maintaining inference-time flexibility.
\item We demonstrate through extensive experiments that LAPO achieves substantial efficiency gains (up to 40.9\% token reduction) while improving accuracy, revealing that models can develop emergent meta-cognitive abilities for reasoning budget allocation.
\end{itemize}

\section{Related Works}
\subsection{Test-Time Scaling in Large Language Models}
Increasing test-time computation has consistently been shown to improve performance in complex reasoning tasks, mathematical problem-solving, and code generation \citep{39,41,30,26}. Test-time scaling laws indicate predictable performance gains from increasing inference computation, either by generating more reasoning chains or longer ones \citep{39,42,27}. Prominent approaches include parallel sampling of multiple reasoning paths \citep{41}, tree-based search \citep{40,39}, and iterative refinement techniques \citep{42,43}. 

Recent reasoning models such as OpenAI’s O1 and DeepSeek’s R1-style models \citep{27,26} simplify test-time scaling by generating extended reasoning traces through reinforcement learning with verifiable rewards (RLVR), encouraging deep thinking behaviors such as broad exploration and feasibility checks \citep{44}. However, these extended reasoning behaviors often lead to much longer reasoning traces, sometimes several times longer than those produced by short CoT models \citep{46,32}, creating an “overthinking” issue that largely increases inference costs \citep{31}. Several recent works have shown that this extended reasoning often includes redundant or unnecessary verification and reflection, even on simple problems \citep{45}. Despite their promising results, existing methods lack precise and dynamic control over the length of generated reasoning chains, resulting in often suboptimal performance or unrealized potential efficiency gains.

\subsection{Efficient Long Chain-of-Thought LLM}
While test-time scaling with long CoT significantly improves accuracy, it comes at the cost of computational inefficiency. In particular, reasoning models often produce verbose and unnecessary reasoning when solving simple problems—a phenomenon commonly referred to as overthinking \citep{46}. To address the overthinking phenomenon in reasoning models, various methods have been proposed following three main strategies. Prompt-based methods attempt to control response length by incorporating instructions directly into prompts \citep{47}, but cannot achieve precise length control. Training-based methods include supervised fine-tuning approaches that collect datasets with variable lengths \citep{11,16,48,49} and RL-based methods that incorporate length penalties into reward functions \citep{6,50,51,52}. However, these methods fail to control length according to users’ requirements or problem complexity. Router-based methods train separate classifiers to route queries between fast and reasoning models \citep{53,54}, but require additional computational overhead. Recent advances in token budget control have introduced more sophisticated approaches. Works like L1 \citep{14} and Elastic Reasoning \citep{52} can more precisely control output length under given token budgets, yet they fail to enable autonomous estimation of appropriate response lengths for different problems.

In contrast to these prior approaches, our LAPO framework uniquely combines autonomous budget estimation and precise length control capabilities through a two-stage reinforcement learning design. Unlike existing methods that rely on external truncation mechanisms or require manual budget specification, LAPO trains models to intrinsically learn appropriate reasoning lengths while maintaining reasoning completeness and logical coherence. This endogenous length control capability enables problem-adaptive token budget allocation, achieving significant efficiency improvements while maintaining or enhancing reasoning performance.

\section{Method}
We present Length-Adaptive Policy Optimization (LAPO), a framework that enables reasoning models to internalize efficient reasoning as an intrinsic capability. Our approach fundamentally differs from existing methods by teaching models to develop an internal understanding of appropriate reasoning depth, rather than imposing external constraints. We achieve this through a carefully designed two-stage training process that first discovers natural reasoning patterns, then transforms these patterns into an internalized capability.

\subsection{Overview}

Consider a reasoning model generating response $r$ for problem $q$. While current models produce high-quality solutions, they lack awareness of computational efficiency, often generating responses far exceeding necessary length. Our goal is to train models that autonomously determine appropriate reasoning lengths while maintaining solution quality.

\begin{figure}[t!]
\centering
\includegraphics[width=\linewidth]{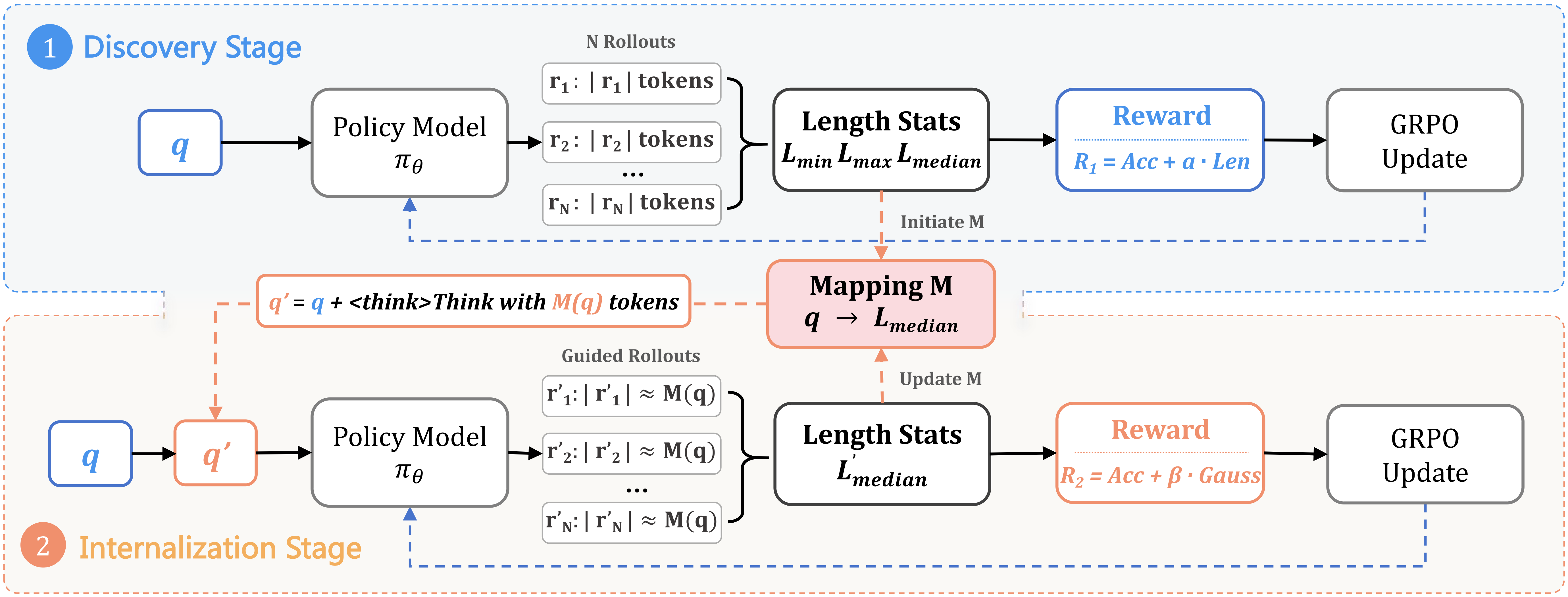}
\caption{The LAPO framework consists of two stages: (1) Discovery stage learns natural reasoning patterns by rewarding efficient correct solutions and collecting length statistics; (2) Internalization stage embeds these statistics as self-proposed plans within the model’s reasoning context, teaching models to internalize efficient reasoning.}
\label{fig}
\end{figure}

Our key insight is that successful problem solutions naturally exhibit certain length distributions that reflect intrinsic problem complexity. Rather than viewing these patterns as constraints to enforce, we treat them as signals that teach models about reasoning depth requirements. LAPO employs a two-stage approach illustrated in Figure \ref{fig}: the Discovery stage explores natural reasoning patterns through length-aware rewards, while the Internalization stage transforms these patterns into adaptive reasoning behavior.

\subsection{Discovery Stage: Learning Natural Reasoning Patterns}

The Discovery stage aims to uncover inherent relationships between problems and their natural reasoning lengths through GRPO training with a carefully designed reward mechanism that encourages efficient exploration while maintaining correctness.

\paragraph{Extracting Statistics from GRPO Rollouts.} During GRPO training, we generate $N$ rollout responses for each problem $q$ in the training batch. From these rollouts, we collect the lengths of responses that produce correct answers:
\begin{equation}
\mathcal{L}_{q} = \{|r_i| : \mathbb{I}(y_i = y_{\text{gold}}) = 1, i \in [1,N]\}
\label{eq-1}
\end{equation}
where $y_i$ is the predicted answer from the $i$-th rollout response $r_i$. This collection, extracted directly from the GRPO sampling process, represents natural variation in successful reasoning lengths.

We derive two key statistics from these rollouts. First, we establish a reasonable length range using percentiles to filter outliers while preserving central tendencies:
\begin{equation}
[L_{\min}, L_{\max}] = [\text{Percentile}_{30}(\mathcal{L}_{q}), \text{Percentile}_{70}(\mathcal{L}_{q})]
\label{eq-2}
\end{equation}
Second, we create a problem-to-length mapping that will guide the Internalization stage:
\begin{equation}
\mathcal{M}: q \mapsto L_{\text{median}}(q) = \text{Median}(\mathcal{L}_{q})
\label{eq-3}
\end{equation}
For problems without correct solutions in the current rollouts, we set $\mathcal{M}(q) = 4096$ (maximum sequence length) to encourage comprehensive exploration in subsequent episodes.

\paragraph{Length-Aware Reward Design.} We employ a composite reward function balancing accuracy and efficiency:
\begin{equation}
R_{D}(r_i, q) = \mathbb{I}(y_i = y_{\text{gold}}) + \alpha \cdot R_{\text{1}}(r_i, q)
\label{eq-4}
\end{equation}

The length component operates on a crucial principle—only correct responses receive length-based rewards. Let $\mathcal{C}_i = \mathbb{I}(y_i = y_{\text{gold}})$ indicate whether the response is correct, and define the distance to the target length range as $d_i = \min(||r_i| - L_{\min}|, ||r_i| - L_{\max}|)$. We introduce a linear decay function $f(d) = \max(0, 1 - d/100)$ to penalize deviations from the efficient length range. The length reward is then defined as:
\begin{equation}
R_{\text{1}}(r_i, q) = 
\begin{cases}
1.0 & \text{if } \mathcal{C}_i = 1 \land |r_i| \in [L_{\min}, L_{\max}] \\
f(d_i) & \text{if } \mathcal{C}_i = 1 \land |r_i| \notin [L_{\min}, L_{\max}] \\
0 & \text{if } \mathcal{C}_i = 0
\end{cases}
\label{eq-5}
\end{equation}

This design creates gradients guiding models toward efficient lengths while allowing flexibility for complex problems. Throughout the Discovery stage, we continuously update $\mathcal{M}$ after each GRPO training step to reflect evolving model capabilities.

\subsection{Internalization Stage: Length-Aware Efficient Reasoning}

The Internalization stage transforms discovered patterns into internalized capabilities through continued GRPO training with modified prompts and rewards.

\paragraph{Length-Conditioned Rollout.} We augment each problem prompt with explicit length guidance:
\begin{tcolorbox}[colback=gray!10, colframe=gray!50, boxrule=0.5pt]
\texttt{prompt}$_q'$ = \texttt{prompt}$_q$ + ``\texttt{<think>} I will answer the question with $n$ tokens.''
\end{tcolorbox}
where $n = \mathcal{M}(q)$ from the Discovery stage. This embeds length awareness within the reasoning context, helping models perceive computational budgets as intrinsic to thinking rather than external constraints.

\paragraph{Length-Adherence Reward.}
To encourage the model to follow its self-declared reasoning budget, the Internalization stage employs a precision-focused reward function. This function is designed to reward the alignment between the model's output length and its self-declared budget n. The total reward is defined as:
\begin{equation}
R_I(r_i, q') = \mathbb{I}(y_i = y_{\text{gold}}) + \beta \cdot R_\text{2}(r_i, q')
\label{eq-7}
\end{equation}
where the adherence component, $R_\text{2}$, is only granted for correct solutions:
\begin{equation}
R_\text{2}(r_i, n) = \begin{cases}
\exp\left(-\frac{(|r_i| - n)^2}{2\sigma^2}\right) & \text{if } \mathcal{C}_i = 1, \\
0 & \text{if } \mathcal{C}_i = 0;
\end{cases}
\label{eq-8}
\end{equation}
This Gaussian-inspired reward positively reinforces solutions that are both correct and consistent with the intended reasoning depth. By rewarding adherence to the self-proposed plan, this mechanism guides the model to internalize the relationship between problem complexity and an appropriate computational budget, rather than merely tracking an external signal.

\begin{algorithm}[t]
\small
\caption{Length-Adaptive Policy Optimization(LAPO)}
\label{alg:lapo}
\begin{algorithmic}[1] %
\State \textbf{Input:} Base model $\pi_\theta$, training data $\mathcal{D}$, hyperparameters $\alpha, \beta, \sigma, E_1, E_2$
\State \textbf{Output:} Length-adaptive model $\pi_\theta^*$
\State
\State \textbf{// Discovery Stage} 
\For{episode $e = 1$ to $E_1$}
    \State Sample batch $\mathcal{B} \subset \mathcal{D}$
    \For{each problem $q \in \mathcal{B}$}
        \State Generate $N$ rollouts: $\{r_1, \dots, r_N\} \sim \pi_\theta(q)$
        \State Collect correct lengths: $\mathcal{L}_q = \{|r_i| : y_i = y_{\text{gold}}\}$
        \State Compute range: $[L_{\min}, L_{\max}] = [\text{P}_{30}(\mathcal{L}_q), \text{P}_{70}(\mathcal{L}_q)]$
        \State Update mapping: $\mathcal{M}(q) = \text{Median}(\mathcal{L}_q)$
        \State $\text{Compute rewards: } R_D(r_i, q) ={} \mathbb{I}(y_i = y_{\text{gold}}) + \alpha \cdot R_{\text{1}}(r_i, q)$
    \EndFor
    \State Update $\pi_\theta$ using GRPO with rewards $R_1$
\EndFor
\State
\State \textbf{// Internalization Stage}
\For{episode $e = 1$ to $E_2$}
    \State Sample batch $\mathcal{B} \subset \mathcal{D}$
    \For{each problem $q \in \mathcal{B}$}
        \State \text{Augment prompt: } $q' \leftarrow{} q + \text{``\texttt{<think>} }\text{I will}$ $\text{ answer the question} \text{ with } \mathcal{M}(q) \text{ tokens.''}$
        \State Generate $N$ rollouts: $\{r_1, \dots, r_N\} \sim \pi_\theta(q')$
        \State $\text{Compute rewards: } R_I(r_i, q') ={} \mathbb{I}(y_i = y_{\text{gold}}) + \beta \cdot R_\text{2}(r_i, q')$
        \State \parbox[t]{\dimexpr\linewidth-2em\relax}{Update mapping $\mathcal{M}(q)$ using dual-strategy (Eq. \ref{eq:update_m})}
    \EndFor
    \State Update $\pi_\theta$ using GRPO with rewards $R_2$
\EndFor
\State \Return $\pi_\theta^*$
\end{algorithmic}
\end{algorithm}

\paragraph{Internalization via In-Context Guidance.} A cornerstone of our framework is how it fosters genuine internalization, enabling inference-time flexibility without explicit length targets. The key lies in the design of the augmented prompt. Placing the self-declarative guidance immediately after the \texttt{<think>} token transforms an external constraint into an intrinsic part of the model's cognitive plan.

During the Internalization stage, we refine $\mathcal{M}$ based on new GRPO rollouts with a dual-strategy update:

\begin{equation}
\mathcal{M}(q) = \begin{cases}
\text{Median}(\mathcal{L}_q^{(t)}) & \text{if previously unsolved} \\
\min(\mathcal{M}(q), \text{Median}(\mathcal{L}_q^{(t)})) & \text{if previously solved}
\end{cases}
\label{eq:update_m}
\end{equation}

This ensures newly solved problems establish reasonable benchmarks while previously solved problems gravitate toward more efficient solutions.

\subsection{Training Pipeline}

We present the complete LAPO training procedure in Algorithm \ref{alg:lapo}. LAPO employs GRPO across both stages with the following pipeline:

\paragraph{Discovery Stage} (Lines 4-15): The model explores natural reasoning patterns through GRPO training with length-aware rewards. For each problem in the training batch, we generate multiple rollouts and extract statistics from successful responses. The mapping $\mathcal{M}$ is continuously updated to capture the evolving understanding of appropriate reasoning lengths. This stage runs for $E_1$ epochs, allowing the model to discover problem-specific length patterns through self-supervised exploration.

\paragraph{Internalization Stage} (Lines 17-27): The model learns to internalize efficient reasoning by incorporating discovered length patterns into the training process. Each problem prompt is augmented with target length information derived from the Discovery stage. The placement of this guidance within the \texttt{<think>} block encourages the model to treat the budget as part of its own reasoning plan, which fosters genuine length awareness rather than rote instruction following. The dual-strategy update mechanism refines the mapping $\mathcal{M}$ throughout training, allowing newly solved problems to establish benchmarks while encouraging efficiency improvements for previously solved ones.

This progressive design mirrors cognitive development: first gaining experience about appropriate reasoning depth through practice, then learning to anticipate these requirements proactively. The embedding of guidance as a self-declared plan is the key mechanism that bridges this gap from experience to proactive anticipation. By making efficiency an intrinsic part of reasoning, LAPO creates models that naturally adapt computational investment to match problem demands.

\section{Experiment Setup}
\section{Experiment Setup}
\paragraph{Training Details.} We train our models on a mixed dataset of 10,000 mathematical problems to ensure a balanced difficulty distribution, comprising 6,000 examples from the DeepScaleR-Preview-Dataset~\cite{21} and 4,000 from the intermediate levels of the MATH dataset~\cite{24}. We apply LAPO to two base models: DeepSeek-R1-1.5B~\cite{26} and DeepScaleR-1.5B-Preview~\cite{21}.

All models are trained using the Group Relative Policy Optimization (GRPO) algorithm. Discovery Stage(LAPO-D) is trained for 3 episodes with reward $R_D$ (Eq. \ref{eq-4}), where hyperparameter $\alpha$ is 0.7. The resulting model then serves as the initialization for the subsequent Internalization Stage(LAPO-I), which is trained for 3 episodes using reward $R_I$ (Eq. \ref{eq-7}) with $\beta$ set to 0.7. These hyperparameters were empirically chosen to provide a strong incentive for efficiency while ensuring correctness remains the primary learning signal. Note that we did not conduct extensive hyperparameter tuning, so one can expect further improvements with additional optimization. Besides, due to computational constraints, the maximum context length is limited to 4,096 tokens during training. Crucially, to ensure a fair comparison, most baselines, including ThinkPrune and L1, were also trained or evaluated under this same 4k context limit. Further hyperparameters are detailed in Appendix.

\paragraph{Evaluation Details.} At inference, we expand the generation window to a generous 32,768 tokens for all models to assess their true, unconstrained reasoning capabilities. This setup allows us to isolate the efficiency gains stemming directly from the LAPO framework, rather than from simple context window limitations. We evaluate on four challenging benchmarks: MATH-500~\cite{24}, AIME2024, AMC23, and Olympiad-Bench~\cite{25}. Following standard practices~\cite{26}, we report both Pass@1 accuracy and the average number of tokens. For each problem, we sample N responses (4 for MATH-500/OlympiadBench, 32 for AIME/AMC) with a temperature of 0.6 and a top-p of 0.95.

\paragraph{Baselines.} We benchmark LAPO against three classes of baselines: the foundational models, an ablation baseline, and existing methods designed for efficient reasoning. First, we evaluate the Base Models to establish a performance starting point. These include DeepSeek-R1-1.5B~\cite{26} and DeepScaleR-1.5B-Preview~\cite{21}. Second, to isolate the effect of our length-reward, we also include an Ablation Baseline, denoted as Acc-Only, which is trained with GRPO using only the accuracy reward. Finally, we compare against several state-of-the-art Efficient Reasoning Baselines, which represent different philosophies for achieving efficiency. (1)Implicit Regularization: HAPO~\cite{57}, which uses history-aware rewards. (2)Budget-Driven Control: L1~\cite{14} and ThinkPrune~\cite{29}, which follow external length targets. (3)Adaptive Activation:  AutoThink~\cite{56}, AdaptThink~\cite{59}, and Thinkless~\cite{58}, which learn a binary think/no-think policy.

\section{Results and Analysis}

\begin{table}[t!]
\centering
\small
\caption{Main results on MATH500, AIME2024, AMC23, and OlympiadBench. We report Pass@1 accuracy (\%) and the average number of generated tokens (\#Tok). For each metric, \textbf{bold} indicates the best and \underline{underline} indicates the second-best Pass@1 score within each base model group.}
\label{tab:t1}
\begin{tabular}{L{2.1cm}C{0.8cm}C{0.6cm}C{0.8cm}C{0.6cm}C{0.8cm}C{0.6cm}C{0.8cm}C{0.6cm}C{0.8cm}C{0.6cm}} 
\toprule
\multirow{2}{*}{} &
  \multicolumn{2}{c}{MATH-500} &
  \multicolumn{2}{c}{AIME2024} &
  \multicolumn{2}{c}{AMC-23} &
  \multicolumn{2}{c}{OlympiadBench} &
  \multicolumn{2}{c}{Average} \\ \cmidrule{2-11} 
                    & Pass@1        & \#Tok & Pass@1        & \#Tok & Pass@1        & \#Tok & Pass@1        & \#Tok & Pass@1        & \#Tok \\ \midrule
\rowcolor{gray!15}
\multicolumn{11}{l}{\textit{Base model: DeepSeek-R1-1.5B}} \\
\midrule
HAPO                & 82.2          & 2288  & \underline{31.3} & 8649  & 67.3          & 4735  & 50.1          & 5024  & 57.7          & 5174  \\
AutoThink           & 83.5          & 2017  & 29.7          & 7084  & 70.2          & 3499  & 51.2          & 4606  & 58.6          & 3825  \\
AdaptThink           & 81.6          & 1580  & 23.9         & 6432  & 63.2          & 2860  & 48.5          & 4616  & 54.3         & 3871  \\
\midrule
Base                & 83.1          & 4031  & 30.3          & 12150 & 68.3          & 7222  & 50.0          & 8942  & 57.9          & 8086  \\
\quad + Acc-Only    & 83.3          & 3061  & \textbf{31.6} & 10628 & 70.5          & 5307  & 50.6          & 6402  & 59.0 & 6349  \\
\quad + LAPO-D      & \textbf{84.7} & 2566  & 28.5          & 8415  & \textbf{72.2} & 4132  & \underline{51.3} & 5595  & \textbf{59.2} & 5177  \\
\quad \quad + LAPO-I      & \underline{84.3}          & 2354  & 29.3          & 8318  & \underline{71.2}          & 3568  & \textbf{51.7} & 4863  & \underline{59.1}          & 4775  \\ 
\midrule
\rowcolor{gray!15}
\multicolumn{11}{l}{\textit{Base model: DeepScaleR-1.5B-Preview}} \\
\midrule
L1-Exact            & 80.6          & 1953  & 24.4          & 2625  & 70.9          & 2177  & 48.8          & 2357  & 56.2          & 2278  \\
L1-Max              & 81.9          & 1673  & 24.9          & 3638  & 72.7          & 2705  & 50.5          & 2151  & 57.5          & 2541  \\
ThinkPrune-I2k   & 85.5          & 1707  & 34.9          & 5095  & 74.3          & 2913  & 54.7          & 3498  & 62.3          & 3303  \\
ThinkPrune-4k       & \textbf{86.6} & 2042  & 35.5          & 6488  & 76.3          & 3839  & 55.7          & 4010  & 63.5          & 4094  \\
HAPO                & 84.4          & 2370  & 31.4          & 7702  & 70.3          & 4301  & 51.4          & 4571  & 59.3          & 4736  \\
AutoThink           & 84.9          & 1635  & 36.2          & 7201  & 67.8          & 3658  & 52.5          & 4085  & 60.4          & 4144  \\ 
Thinkless           & 81.3          & 2944  & 28.9          & 9143  & 65.7          & 5276  & 50.2          & 6057  & 56.5          & 5855  \\ 
\midrule
Base                & 85.8          & 3280  & 35.5          & 9246  & 74.2          & 6416  & 54.6          & 5974  & 62.5          & 6229  \\
\quad + Acc-Only    & 85.6          & 2510  & 36.9          & 7319  & \underline{77.6}          & 4244  & 55.6          & 4712  & 63.9          & 4696  \\
\quad + LAPO-D      & \underline{86.4} & 2365  & \underline{37.6}          & 5945  & \underline{77.6}          & 3655  & \underline{56.1}          & 4499  & \underline{64.4}          & 4116  \\
\quad \quad + LAPO-I      & 86.3          & 2168  & \textbf{38.1} & 5371  & \textbf{78.3} & 3765  & \textbf{56.3} & 4024  & \textbf{64.8} & 3832  \\ 
\bottomrule
\end{tabular}
\end{table}

We present comprehensive experimental results and analysis to validate LAPO’s effectiveness and understand its underlying mechanisms. We begin with the main results (Section~\ref{sec:main_results}), benchmarking LAPO against baselines and state-of-the-art methods. We then conduct in-depth ablation studies on key design choices, including the the form of length guidance (Section~\ref{sec:ablation_length_guidance}) and the statistical metrics for target length selection (Section~\ref{sec:ablation_statistical}). And a targeted experiment demonstrating the model's robust internalization of reasoning efficiency (Section~\ref{sec:internalization_analysis}). Finally, we provide a mechanistic analysis of how LAPO works, examining its emergent ability for difficulty-aware resource allocation (Section~\ref{sec:analysis_difficulty_aware_allocation}), its qualitative refinement of reasoning patterns (Section~\ref{sec:analysis_reasoning_behaviors}).

\subsection{Main Results}
\label{sec:main_results}
As shown in Table \ref{tab:t1}, LAPO achieves a superior balance of reasoning accuracy and computational efficiency, consistently outperforming its base models and establishing a new state-of-the-art frontier among methods that do not rely on external length controls.

\paragraph{LAPO simultaneously enhances reasoning performance and reduces test-time computes.}
Compared to its base models, LAPO delivers substantial gains. On DeepScaleR-1.5B-Preview, it reduces tokens by 38.5\% while boosting average accuracy by 2.3 points; a similar trend holds for DeepSeek-R1-1.5B (41.0\% token cut and 1.2 point accuracy gain). This validates that LAPO learns to produce more concise yet effective reasoning.

\paragraph{LAPO surpasses existing efficient reasoning optimization approaches.} When compared with leading efficiency methods, LAPO consistently demonstrates a superior accuracy-efficiency trade-off. On the more capable DeepScaleR-1.5B base model, LAPO-I achieves the highest average accuracy among all tested methods. This advantage holds across different baseline paradigms. It surpasses budget-driven methods like ThinkPrune-4k and L1-Max under a fair 4k training context. Compared to implicit regularization methods like HAPO, LAPO shows a clear advantage in preserving accuracy. Furthermore, while adaptive activation methods like AutoThink can be highly token-efficient, they do not reach the same level of reasoning quality. This comprehensive comparison highlights that LAPO's ``Discover-Internalize'' process, which fosters an autonomous and continuous length adaptation, leads to a more robust and effective reasoning policy than methods relying on external budgets, progressive compression, or binary mode-switching.

\paragraph{Both Discovery and Internalization stages contribute to the final performance.} LAPO-D first establishes a strong foundation, achieving a significant 36.0\% token reduction on its own by learning natural reasoning length distributions. This is highlighted by comparing it to the Acc-Only baseline. While simply finetuning for accuracy yields some token reduction, LAPO-D's length-aware reward achieves substantially greater efficiency while also improving average accuracy by 0.5 points. This demonstrates that encouraging conciseness via our reward not only prunes redundant thoughts but also helps the model find more robust reasoning patterns. Building on this superior foundation, LAPO-I achieves an additional 6.9\% efficiency gain by internalizing these patterns through in-context guidance. This progressive refinement indicates that our framework learns a generalizable principle of adaptive reasoning.

\subsection{Ablation Study on In-Context Guidance}
\label{sec:ablation_length_guidance}
To validate that our method's success stems from internalizing a self-proposed plan, we ablate the two key factors of our in-context guidance: its form (how precise the guidance is) and its position (whether it's part of the model's internal thought process). We compare our default approach (w/ Exact) against three variants: w/ Range (less precise guidance), w/ Outside (placing the guidance before \texttt{<think>}), and w/ Implicit (no guidance, relying only on the reward). As shown in Table \ref{tab:t2}, the results demonstrate that both form and position are critical for effective internalization.

Our default method outperforms the less precise Range variant, indicating that specific targets discovered in Discovery stage provide a stronger learning signal. More critically, the guidance's position determines whether the model internalizes a plan or merely follows instructions. Moving the guidance outside the \texttt{<think>} block transforms it into an external command and causes accuracy to drop significantly to 63.9\%. This illustrates that the model performs best when the budget is framed as part of its own cognitive plan. Finally, removing the guidance entirely results in the worst performance, with accuracy dropping to 63.6\% and token count reverting to the LAPO-D baseline. This indicates that our explicit, properly-positioned, self-declarative guidance is the critical mechanism for internalization.

\begin{table}[t!]
\centering
\small
\caption{Experimental results with different length guidance for LAPO-I. \textbf{Bold} and \underline{underline} indicate the best and second-best Pass@1 scores. w/ indicates the length guidance used in LAPO-I.}
\label{tab:t2}
\begin{tabular}{L{2.5cm}C{0.8cm}C{0.6cm}C{0.8cm}C{0.6cm}C{0.8cm}C{0.6cm}C{0.8cm}C{0.6cm}C{0.8cm}C{0.6cm}}
\toprule
\multirow{2}{*}{\textbf{Method}} &
  \multicolumn{2}{c}{MATH-500} &
  \multicolumn{2}{c}{AIME2024} &
  \multicolumn{2}{c}{AMC-23} &
  \multicolumn{2}{c}{OlympiadBench} &
  \multicolumn{2}{c}{Average} \\ \cmidrule{2-11} 
                    & Pass@1        & \#Tok & Pass@1        & \#Tok & Pass@1        & \#Tok & Pass@1        & \#Tok & Pass@1        & \#Tok \\ \midrule
\rowcolor{gray!20}
\multicolumn{11}{l}{\textit{Base model: DeepScaleR-1.5B-Preview}}\\ \midrule
Base & 85.8 & 3280 & 35.5 & 9246 & 74.2 & 6416 & 54.6 & 5974 & 62.5 & 6229 \\
\quad LAPO-D & 86.4 & 2365 & \underline{37.6} & 5945 & \underline{77.6} & 3655 & 56.1 & 4499 & \underline{64.4} & 4116 \\ 
\quad\quad w/ Exact & 86.3 & 2168 & \textbf{38.1} & 5371 & \textbf{78.3} & 3765 & \textbf{56.3} & 4024 & \textbf{64.8} & 3832 \\
\quad\quad w/ Range & \underline{86.6} & 2153 & 36.5 & 6095 & 76.9 & 3600 & \underline{56.2} & 4011 & 64.1 & 3964 \\
\quad\quad w/ Outside & 86.5 & 2251 & 36.4 & 5882 & 76.3 & 3850 & 55.4 & 4105 & 63.9 & 4022 \\
\quad\quad w/ Implicit   & \textbf{86.9} & 2181 & 36.2 & 5963 & 76.1 & 4002 & 55.1 & 4206 & 63.6 & 4088 \\
\bottomrule
\end{tabular}%
\end{table}

\begin{table}[t!]
\centering
\small
\caption{Experimental results within different statistical metrics used for target length selection in LAPO-I. \textbf{Bold} and \underline{underline} indicate the best and second-best Pass@1 scores. w/ indicates statistical metrics used for target length selection in LAPO-I.}
\label{tab:t3}
\begin{tabular}{L{2.3cm}C{0.8cm}C{0.6cm}C{0.8cm}C{0.6cm}C{0.8cm}C{0.6cm}C{0.8cm}C{0.6cm}C{0.8cm}C{0.6cm}}
\toprule
\multirow{2}{*}{\textbf{Method}} &
  \multicolumn{2}{c}{MATH-500} &
  \multicolumn{2}{c}{AIME2024} &
  \multicolumn{2}{c}{AMC-23} &
  \multicolumn{2}{c}{OlympiadBench} &
  \multicolumn{2}{c}{Average} \\ \cmidrule{2-11} 
                    & Pass@1        & \#Tok & Pass@1        & \#Tok & Pass@1        & \#Tok & Pass@1        & \#Tok & Pass@1        & \#Tok \\ \midrule
\rowcolor{gray!20}
\multicolumn{11}{l}{\textit{Base model: DeepScaleR-1.5B-Preview}}\\ \midrule
Base & 85.8 & 3280 & 35.5 & 9246 & 74.2 & 6416 & 54.6 & 5974 & 62.5 & 6229 \\
\quad LAPO-D & \textbf{86.4} & 2365 & \underline{37.6} & 5945 & \underline{77.6} & 3655 & 56.1 & 4499 & \underline{64.4} & 4116 \\
\quad\quad w/ Median & \underline{86.3} & 2168 & \textbf{38.1} & 5371 & \textbf{78.3} & 3765 & \underline{56.3} & 4024 & \textbf{64.8} & 3832 \\
\quad\quad w/ Mean       & 85.6 & 2308 & 36.8 & 6030 & 77.4 & 3658 & \textbf{56.6} & 4164 & 64.1 & 4040 \\
\quad\quad w/ Minimum    & 85.9 & 2031 & 36.3 & 6080 & 76.7 & 3324 & 55.0 & 3851 & 63.5 & 3821 \\
\bottomrule
\end{tabular}
\end{table}

\subsection{Ablation on Statistical Metrics for Target Length}
\label{sec:ablation_statistical}
The choice of a statistical measure to derive the target length \(n\) from the distribution of successful solutions is critical. We conduct an ablation study comparing three strategies for this selection: using the median (our default), the mean, and the minimum length.

As shown in Table \ref{tab:t3}, the median proves to be the most effective choice, achieving the best balance between accuracy and efficiency. Using the median as the target yields the highest average accuracy (64.8\%) with an efficient token count of 3,832. This validates our hypothesis that the median, due to its robustness to outliers, provides the most representative signal of a “typically” effective reasoning depth. In contrast, the mean is susceptible to a few excessively long, successful solutions, leading it to set overly generous budgets, resulting in higher token usage (4,040) and slightly lower accuracy. The minimum, while achieving the most aggressive compression (3,821 tokens), suffers a significant accuracy drop (to 63.5\%), suggesting it promotes an over-shortening strategy that discards necessary reasoning steps. These findings underscore the importance of robust statistical measures for learning a well-calibrated reasoning-efficiency trade-off.

\begin{wraptable}{r}{0.5\textwidth}
\centering
\caption{Robustness of LAPO-I to conflicting length instructions on MATH-500.}
\label{tab:robustness_final}
\small
\begin{tabular}{l c cc}
\toprule
\multirow{2}{*}{\textbf{Method}} & \multirow{2}{*}{\textbf{\makecell{Length \\ Constraint}}} & \multicolumn{2}{c}{\textbf{MATH-500}} \\ 
\cmidrule(lr){3-4}
& & Pass@1 (\%) & \#Tok \\ 
\midrule
\rowcolor{gray!20}
\multicolumn{4}{l}{\textit{LAPO-I}}\\ \midrule
Base & N/A & \textbf{86.3} & 2168 \\
\quad +Short & 500 & \underline{86.0} & 2279 \\
\quad +Long & 3500 & 85.9 & 2300 \\
\midrule
\rowcolor{gray!20}
\multicolumn{4}{l}{\textit{LAPO-I w/ Outside}}\\ \midrule
Base & N/A & \textbf{86.2} & 2251 \\
\quad +Short & 500 & 85.1 & 1247 \\
\quad +Long & 3500 & \underline{86.1} & 2821 \\
\bottomrule
\end{tabular}
\end{wraptable}

\subsection{Analysis of Internalization}
\label{sec:internalization_analysis}
To validate that LAPO fosters genuine internalization, we stress-tested our default LAPO-I model against the w/ Outside ablation variant using adversarial Short (500 tokens) and Long (3500 tokens) length prompts. The results in Table \ref{tab:robustness_final} reveal a stark behavioral divergence. Our default LAPO-I remains robust, its output length staying stable around its 2200-token baseline, thus ignoring the conflicting external instructions. In contrast, the w/ Outside model is clearly influenced: its token count drops to 1247 under the Short constraint and rises to 2821 under the Long one. This comparison indicates that the placement of guidance is critical. Framing the budget as part of the model's internal plan (inside \texttt{<think>}) builds a robust, internalized behavior. Framing it externally teaches superficial instruction-following. This indicates the observed robustness of LAPO-I is a direct result of our internalization mechanism.

\subsection{Difficulty-Aware Computational Allocation}
\label{sec:analysis_difficulty_aware_allocation}
\begin{figure}[t!] %
    \begin{minipage}[b]{0.49\textwidth} %
        \centering
        \includegraphics[width=\linewidth]{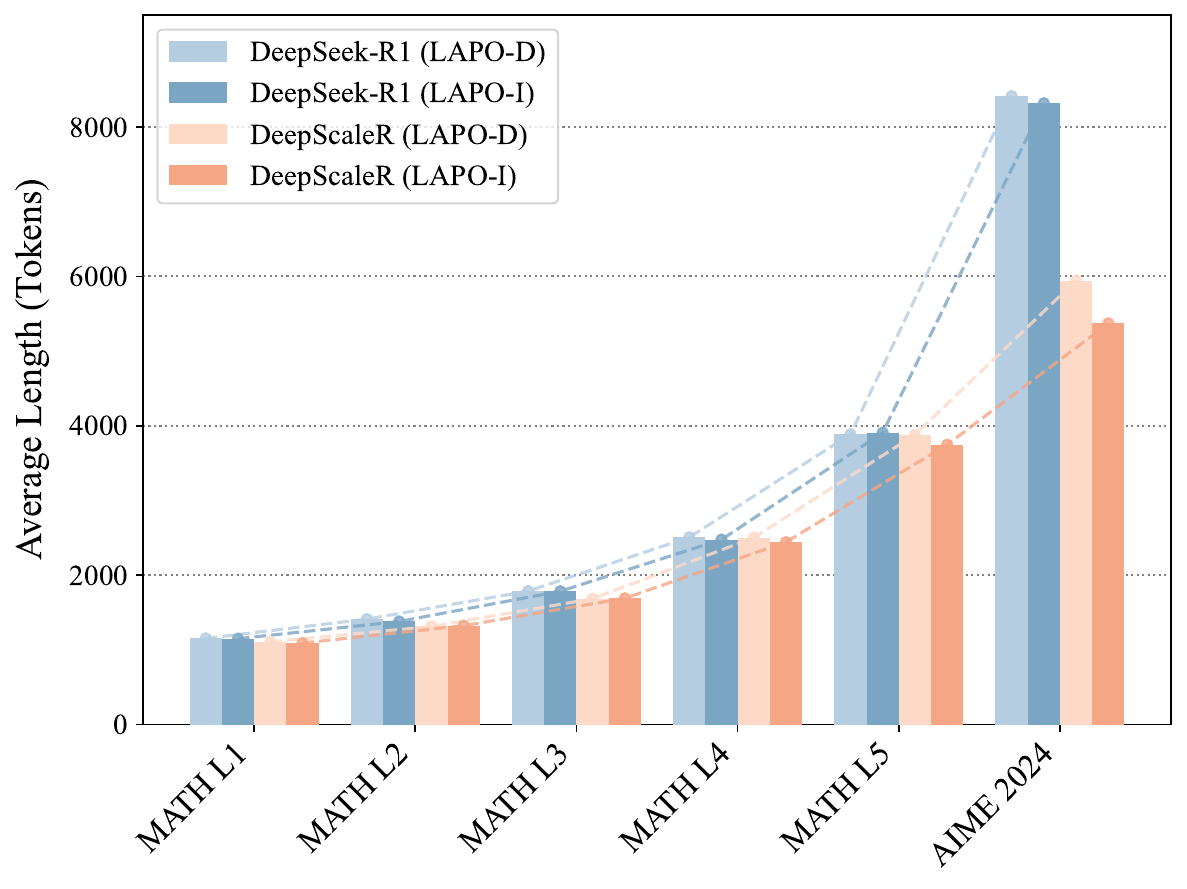} %
        \caption{Reasoning length allocation across mathematical problem difficulty levels. LAPO learns to scale computation with complexity.}
        \label{fig-3}
    \end{minipage}
    \hfill %
    \begin{minipage}[b]{0.49\textwidth} %
        \centering
        \includegraphics[width=\linewidth]{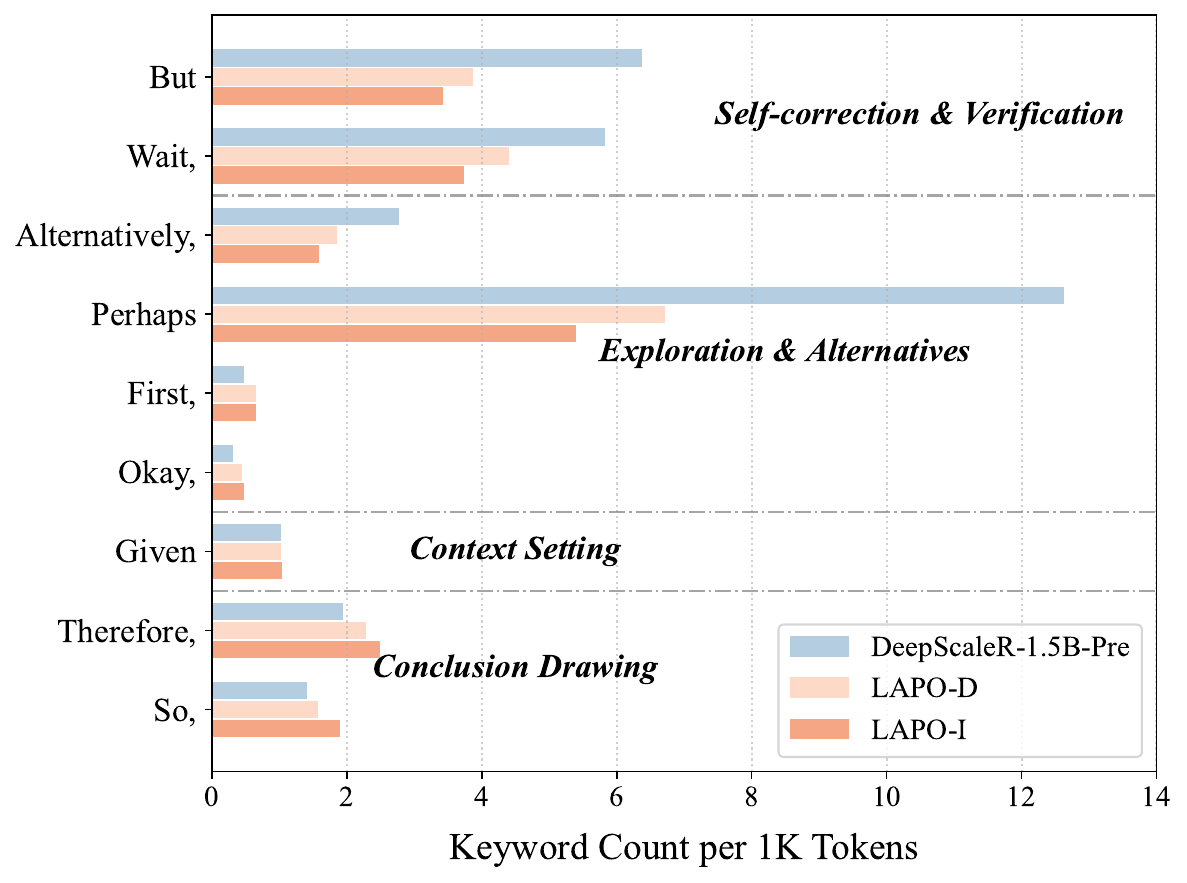}
        \caption{Keyword usage of reasoning behaviors across different stages. LAPO selectively prunes hesitant and exploratory thought patterns.}
        \label{fig-6}
    \end{minipage}
\end{figure}

To understand the mechanisms behind LAPO's efficiency gains, we examine its ability to allocate computational resources in proportion to problem complexity. We evaluate LAPO-trained models on benchmarks with clear difficulty gradients, from MATH Level 1 up to the highly complex AIME 2024. As shown in Figure \ref{fig-3}, our models demonstrate a remarkable emergent capability for difficulty-aware resource allocation. There is a clear, near-linear positive correlation between problem complexity and the average reasoning length. On simpler problems, the models generate concise responses, while for the most challenging AIME questions, they produce extensive reasoning chains that are substantially longer than any solution observed during the training phase. This ability to extrapolate reasoning depth well beyond the bounds of their training experience is a crucial finding. It provides strong evidence that LAPO does not merely teach models to compress their outputs. Instead, it successfully imparts a generalizable principle of complexity-to-length mapping. This allows the models to dynamically and appropriately scale their computational investment when faced with novel problems of varying difficulty. The consistent scaling behavior across different base models further underscores that LAPO develops a robust, fundamental reasoning strategy rather than model-specific optimizations.

\subsection{Qualitative Refinement of Reasoning Behaviors}
\label{sec:analysis_reasoning_behaviors}
Beyond quantitatively adjusting how much computation the model uses, we next investigate how it qualitatively refines the reasoning process to achieve this efficiency. To do this, we analyzed the frequency of keywords indicative of different cognitive behaviors, such as ``Self-Correction and Verification'', ``Exploration and Alternatives'', ``Context Setting'', and ``Conclusion Drawing'', in the generated responses (Figure \ref{fig-6}). The results reveal a significant shift in the model’s reasoning style. The most significant change is a dramatic reduction in keywords associated with Self-Correction and Exploration. The base model (DeepScaleR-1.5B-Pre) frequently employs these terms, indicating a verbose and deliberative reasoning style. After LAPO-D training, and further refined in LAPO-I, the model significantly curtails this internal monologue. This suggests that LAPO effectively discourages redundant verification loops and inefficient exploration of the solution space. Crucially, this efficiency gain is not achieved by indiscriminately shortening the response. The frequency of keywords related to Context Setting and Conclusion Drawing remains stable across training stages. This demonstrates that LAPO selectively prunes inefficient and hesitant thought patterns while preserving the essential scaffolding of a coherent logical argument. The model learns to maintain its ability to properly frame a problem and articulate its deductions.

\section{Conclusion}

In this work, we introduce Length-Adaptive Policy Optimization (LAPO), a two-stage reinforcement learning framework that enables language models to autonomously adjust reasoning length based on problem complexity.  Unlike existing approaches that impose uniform constraints, LAPO recognizes that efficient reasoning requires understanding problem-specific computational needs rather than following rigid rules. Our two-stage design enables a natural progression: models first learn what constitutes appropriate reasoning depth through experience, then develop the ability to anticipate these requirements proactively.  This approach mirrors how human experts develop intuition about problem complexity, allocating mental effort proportionally to task demands. Extensive experiments validate LAPO's effectiveness, achieving remarkable efficiency gains with up to 40.9\% reduction in token usage while simultaneously improving accuracy by 2.3\% on mathematical reasoning benchmarks.  Our analysis reveals that this improvement stems from the model's ability to distinguish between problems requiring elaborate derivations versus those needing only brief calculations.  These results confirm that when models learn from their own successful patterns rather than arbitrary constraints, they develop more robust and efficient reasoning strategies.

\bibliographystyle{iclr2026_conference}
\bibliography{ref}

\appendix
\newpage
\section{Technical Appendices and Supplementary Material}

\subsection{Implementation Details}
\label{appendix:implementation_details}

\textbf{System prompt used for training.} 
The system prompts used for the two-stage training are shown in the boxes below. The prompt titled \textbf{LAPO-D-prompt} was used for DeepSeek-R1-Distill-Qwen-1.5B, and \textbf{LAPO-I-prompt} was used for DeepScaleR. This approach maintains consistency with the original RL training of DeepSeek-R1.

\begin{AIbox}{LAPO-D-prompt}
You are a helpful assistant. A conversation between User and Assistant. The user asks a question, and the Assistant solves it. The Assistant first thinks about the reasoning process in the mind and then provides the user with the answer. The reasoning process is enclosed within \textless think\textgreater{} and \textless/think\textgreater{} tags, respectively, i.e., \textless think\textgreater{} reasoning process here \textless/think\textgreater{} answer here. User: \{question\} Please think step by step and output the final answer within \texttt{\textbackslash boxed\{\}}. Assistant: \textless think\textgreater
\end{AIbox}

\begin{AIbox}{LAPO-I-prompt}
You are a helpful assistant. A conversation between User and Assistant. The user asks a question, and the Assistant solves it. The Assistant first thinks about the reasoning process in the mind and then provides the user with the answer. The reasoning process is enclosed within \textless think\textgreater{} and \textless/think\textgreater{} tags, respectively, i.e., \textless think\textgreater{} reasoning process here \textless/think\textgreater{} answer here. User: \{question\} Please think step by step and output the final answer within \texttt{\textbackslash boxed\{\}}. Assistant: \textless think\textgreater{} I will answer the question with \{length\} tokens.
\end{AIbox}

\textbf{Training and Reproduction Details.} We trained the model on the OpenRLHF framework. During training, we sampled 8 responses for each query in the batch with a temperature of 1.0, set the kl parameter to 0.0001, used a learning rate of 1e-6 and a batch size of 128, and set the maximum context length to 4K tokens during training. Both LAPO-D and LAPO-I training were conducted for 3 episodes, approximately 240 steps. The \(\alpha\) and \(\beta\) parameters in \(R_1\) and \(R_2\) were 0.7 and 0.8, respectively. All experiments were conducted using 4 A800 GPUs. We provide training hyperparameters in Table \ref{tab:t7}.

\begin{wraptable}{r}{0.4\textwidth}

\caption{Training Hyperparameters}
\centering
\label{tab:t7}
\begin{tabular}{ll}
\toprule
\textbf{Hyperparameter}        & \textbf{Value}  \\ \midrule
Epochs                & 1      \\
Episodes              & 3      \\
Learning Rate         & 1e-6   \\
Train Batch Size      & 128    \\
Temperature           & 1.0    \\
Rollout per Prompt    & 8      \\
Prompt Max Length     & 1024   \\
Generation Max Length & 4096   \\
KL Coefficient        & 0.0001 \\
Precision             & BF16   \\
\(\alpha\)            & 0.7    \\
\(\beta\)             & 0.7    \\ \bottomrule
\end{tabular}
\end{wraptable}

\subsection{Training Dynamics}
\label{sec:training_dynamics}
\begin{figure*}[t]
    \centering
    \begin{subfigure}[b]{0.49\textwidth}
        \centering
        \includegraphics[width=\linewidth]{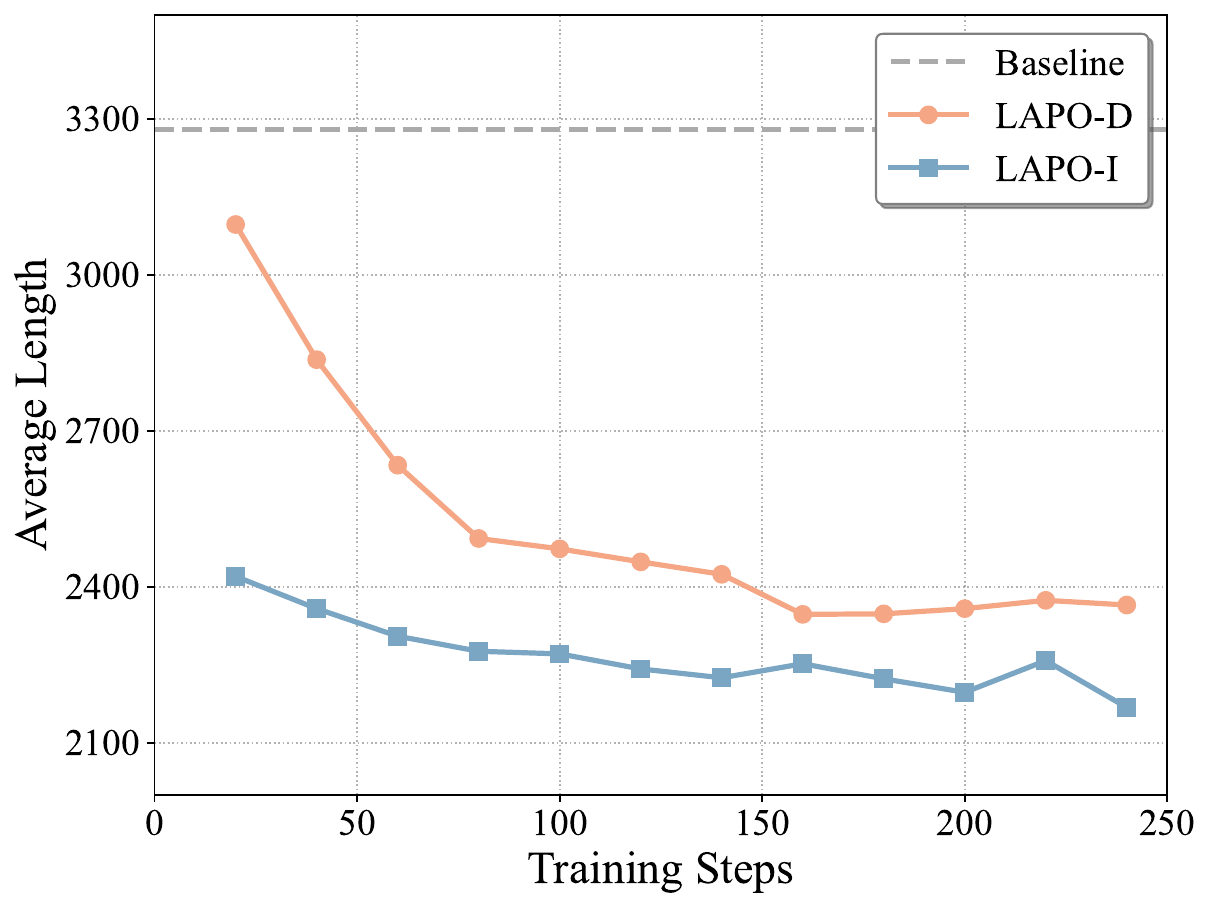}
        \caption{Average Length on MATH-500}
        \label{fig:len_dynamics}
    \end{subfigure}
    \hfill %
    \begin{subfigure}[b]{0.49\textwidth}
        \centering
        \includegraphics[width=\linewidth]{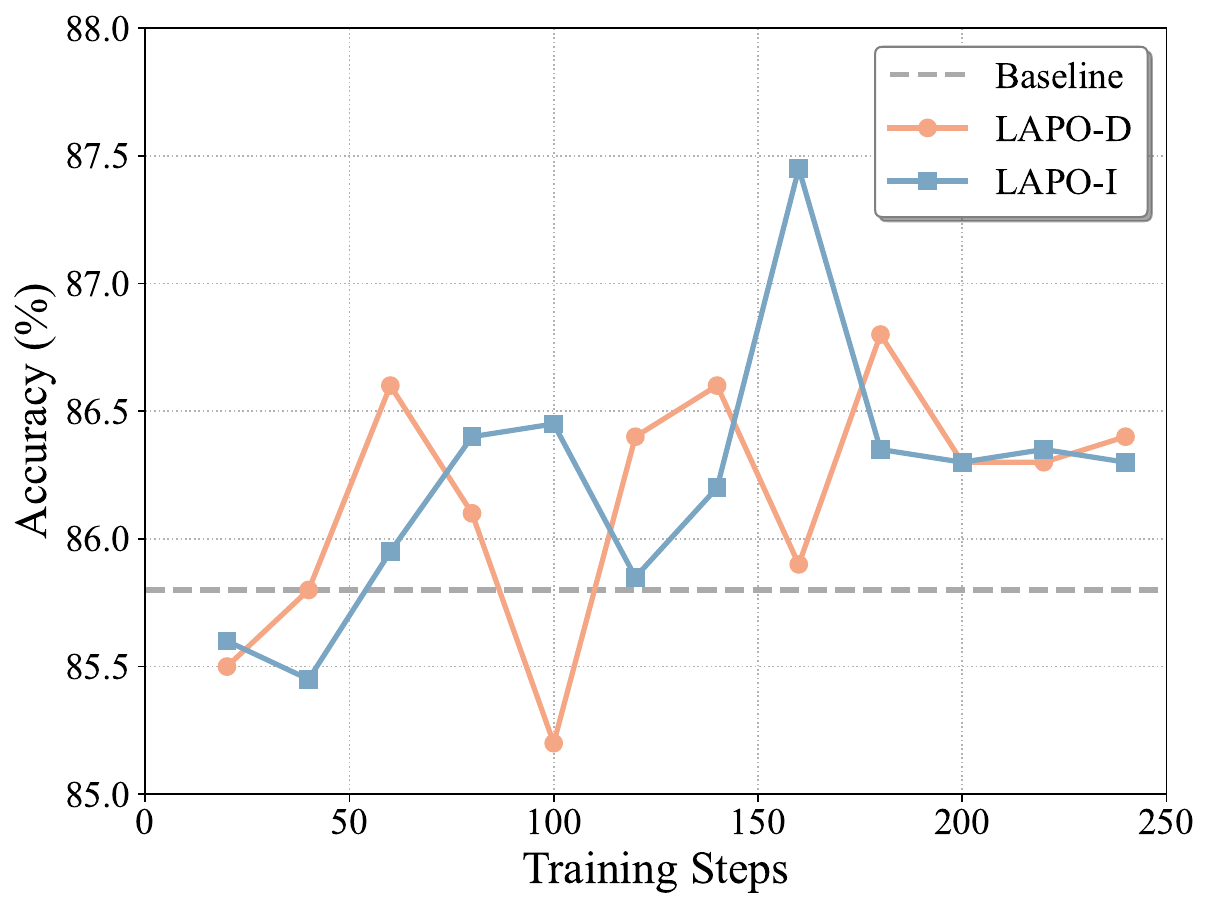}
        \caption{Accuracy on MATH-500}
        \label{fig:acc_dynamics}
    \end{subfigure}
    \caption{Training dynamics evaluated on the MATH-500 validation set. Checkpoints were saved periodically during training on our mixed dataset. (a) Both LAPO-D and LAPO-I policies learn to significantly reduce the average response length. (b) These efficiency gains are achieved while maintaining or even improving accuracy over the baseline.}
    \label{fig:training_dynamics}
\end{figure*}

We analyze the training dynamics by periodically evaluating model checkpoints on the MATH-500 validation set to understand the learning mechanisms of our two-stage framework. As illustrated in Figures \ref{fig:len_dynamics} and \ref{fig:acc_dynamics}, LAPO achieves a superior balance between efficiency and accuracy across both training stages.

\textbf{Continuous Efficiency Gains.} Figure \ref{fig:len_dynamics} shows a clear, two-step reduction in token generation. In Stage 1, the LAPO-D policy rapidly becomes more concise, with its average length decreasing from a verbose baseline of ~3,280 tokens to a stable ~2,365 tokens, driven by the length-aware reward ($R_1$). Building on this, the LAPO-I policy achieves further compression, reducing the length to below 2,200 tokens. This demonstrates that the plan-adherence reward ($R_2$), combined with in-context guidance, effectively encourages the model to execute its self-proposed reasoning plans more precisely.

\textbf{Accuracy Maintenance and Refinement.} Crucially, these efficiency gains do not compromise performance. As shown in Figure \ref{fig:acc_dynamics}, accuracy on MATH-500 is consistently maintained or improved. The LAPO-D policy's accuracy climbs from 85.8\% to over 86.4\%, suggesting the reward mechanism prunes redundant or error-prone reasoning steps. The LAPO-I policy sustains this high accuracy level even on a much tighter token budget. Notably, it exhibits a transient performance peak, a key finding that suggests the in-context guidance actively steers the model toward more focused and effective reasoning, rather than merely acting as a constraint.

In summary, the training dynamics validate our two-stage design. LAPO-D establishes a robust foundation for efficient reasoning, which LAPO-I then refines to achieve a superior performance-cost balance. The smooth convergence on a challenging validation set confirms that by learning from its own successful patterns, the model develops transferable and efficient reasoning strategies.

\begin{table}[t!]
\small
\centering
\caption{Ablation study on the training dataset. This table compares performance when trained on different data sources. For each metric column, \textbf{bold} indicates the best score and \underline{underline} indicates the second-best score across all configurations.}
\label{tab:t6}
\begin{tabular}{L{2cm}C{0.8cm}C{0.6cm}C{0.8cm}C{0.6cm}C{0.8cm}C{0.6cm}C{0.8cm}C{0.6cm}C{0.8cm}C{0.6cm}}
\toprule
\multirow{2}{*}{\textbf{Method}} &
  \multicolumn{2}{c}{MATH500} &
  \multicolumn{2}{c}{AIME2024} &
  \multicolumn{2}{c}{AMC-23} &
  \multicolumn{2}{c}{OlympiadBench} &
  \multicolumn{2}{c}{Average} \\ \cmidrule{2-11} 
              & Pass@1        & \#Tok & Pass@1        & \#Tok & Pass@1        & \#Tok & Pass@1        & \#Tok & Pass@1        & \#Tok \\ \midrule
\rowcolor{gray!15}
\multicolumn{11}{l}{\textit{Training Data: Combined (Ours)}} \\ \midrule
\quad LAPO-D        & 86.4 & 2365 & 37.6 & 5945 & 77.6 & 3655 & 56.1 & 4499 & \underline{64.4} & 4116     \\
\quad LAPO-I        & 86.3 & 2168 & \textbf{38.1} & 5371 & \textbf{78.3} & 3765 & \textbf{56.3} & 4024 & \textbf{64.8} & 3832     \\ \midrule
\rowcolor{gray!15}
\multicolumn{11}{l}{\textit{Training Data: DeepScaleR-only}} \\ \midrule
\quad LAPO-D        & 86.1 & 2397 & 36.8 & 6153 & 76.8 & 3983 & 55.5 & 4258 & 63.8 & 4197 \\
\quad LAPO-I        & 86.1 & 2210 & 36.5 & 6418 & 77.0 & 3791 & 55.6 & 3933 & 63.8 & 4088 \\ \midrule
\rowcolor{gray!15}
\multicolumn{11}{l}{\textit{Training Data: MATH-only}} \\ \midrule
\quad LAPO-D        & \textbf{86.5} & 2398 & \underline{38.0} & 7034 & \underline{77.3} & 4060 & \underline{55.8} & 4494 & \underline{64.4} & 4496     \\
\quad LAPO-I        & \underline{86.1} & 2340 & 35.5 & 6452 & 75.8 & 4021 & 54.5 & 4194 & 63.0 & 4251     \\ \bottomrule
\end{tabular}
\end{table}

\subsection{Selection of Training Dataset}
As mentioned in section 4 Experiment Setup, we chose a mixed dataset for training in our experiments. In this section, we provide a detailed analysis of the impact of different dataset selections on model performance. Table \ref{tab:t6} shows the test results on various benchmarks after two-stage training using different training datasets. Several important findings can be observed from the experimental results. Combined-data achieved the best performance in terms of average accuracy, showing a clear advantage over single-dataset training. This indicates that a dataset with a balanced difficulty distribution helps enhance the model's generalization ability across different types of questions. In terms of token usage efficiency, the model trained on combined-data also performed the best. This suggests that problems with different difficulty gradients help establish a more accurate complexity-length mapping relationship. By exposing the model to a wider range of problem difficulties, it can better learn the optimal thinking range for different questions. Taking all these factors into consideration, we selected the mixed dataset as the training data to expose the model to a more diverse set of problems and enable it to deeply learn the optimal reasoning patterns for different questions.

\begin{wraptable}{r}{0.5\textwidth}
\centering
\caption{Performance on the GPQA benchmark. LAPO demonstrates generalizable efficiency and accuracy gains in a non-mathematical, knowledge-intensive domain.}
\label{tab:gpqa_results}
\begin{tabular}{l cc}
\toprule
\textbf{Method} & \textbf{Pass@1 (\%)} & \textbf{\#Tokens} \\
\midrule
\rowcolor{gray!20}
\multicolumn{3}{l}{\textit{Base Model: DeepSeek-R1-1.5B}} \\ \midrule
Base & 36.1 & 10297 \\
\quad + LAPO-D & \textbf{38.1} & 7596 \\
\quad + LAPO-I & \underline{36.9} & 7235 \\
\midrule
\rowcolor{gray!20}
\multicolumn{3}{l}{\textit{Base Model: DeepScaleR-1.5B-Preview}} \\ \midrule
Base & 36.1 & 7667 \\
\quad + LAPO-D & \textbf{38.3} & 6176 \\
\quad + LAPO-I & \underline{37.8} & 6154 \\
\bottomrule
\end{tabular}
\end{wraptable}

\subsection{Generalizability to Expert-Level Question Answering.}
To test if LAPO's benefits extend beyond structured mathematical reasoning, we evaluated our method on the GPQA benchmark. The results, presented in Table \ref{tab:gpqa_results}, demonstrate that LAPO's core principles are highly generalizable.

For both base models, LAPO achieves a compelling dual improvement in accuracy and efficiency. On the DeepSeek-R1-1.5B model, LAPO-D improves Pass@1 accuracy by a significant 2.0 points while reducing token generation by 26.2\%. Similarly, on the more advanced DeepScaleR-1.5B-Preview, LAPO-D boosts accuracy by 2.2 points and cuts tokens by 19.4\%. The internalization stage consistently pushes efficiency further while maintaining a strong accuracy improvement over the baseline. This robust performance on a knowledge-intensive, non-mathematical task indicates that LAPO is not merely exploiting domain-specific patterns. Instead, it learns a fundamental and transferable skill: how to allocate cognitive effort efficiently for complex reasoning across different domains.

\end{document}